\documentclass[runningheads]{llncs}
\usepackage{graphicx}
\usepackage{placeins}
\usepackage{subcaption}

\begin{document}
\title{Label noise in segmentation networks : mitigation must deal with bias}
\titlerunning{Label noise in segmentation networks}
%
\author{Eugene Vorontsov \and
Samuel Kadoury}
\authorrunning{E. Vorontsov et al.}
%
\institute{École Polytechnique de Montréal, Montréal QC H3T1J4, Canada
\email{eugene.vorontsov@gmail.com}}
\maketitle              
\begin{abstract}
Imperfect labels limit the quality of predictions learned by deep neural networks. This is particularly relevant in medical image segmentation, where reference annotations are difficult to collect and vary significantly even across expert annotators. Prior work on mitigating label noise focused on simple models of mostly uniform noise. In this work, we explore biased and unbiased errors artificially introduced to brain tumour annotations on MRI data. We found that supervised and semi-supervised segmentation methods are robust or fairly robust to unbiased errors but sensitive to biased errors. It is therefore important to identify the sorts of errors expected in medical image labels and especially mitigate the biased errors.
\keywords{Label noise  \and Segmentation \and Neural networks}
\end{abstract}
\section{Introduction}

The reference annotations used to train neural networks for the segmentation of medical images are few and imperfect. The number of images that can be annotated is limited by the need for expert annotators and the result is subject to high inter- and intra-annotator variability~\cite{vorontsov2019deep}. Furthermore, the objects targetted by medical image segmentation also tend to be highly variable in appearance. Thus, to make the best of use of limited labeled data, it is important to understand which sorts of errors in reference annotations most affect the segmentation performance of deep neural networks.

Noisy labels can be dealt with by modeling the noise~\cite{patrini2017making,sukhbaatar2014training,mnih2012learning}, re-weighting the contribution of labels depending on some estimate of their reliability~\cite{karimi2020deep,mirikharaji2019learning}, training on pseudo-labels~\cite{karimi2020deep,nie2018asdnet}, designing noise-tolerant objective functions~\cite{wang2020noise,ghosh2017robust}, or estimating true labels~\cite{dawid1979maximum,whitehill2009whose,welinder2010multidimensional,zhou2015regularized,khetan2017learning}.

A generative model of the noise was presented in~\cite{mnih2012learning}. A true segmentation map is estimated using this model, and the segmentation model is updated accordingly. For this approach, a good estimate of the noise model must be known. In \cite{sukhbaatar2014training} and \cite{patrini2017making}, it is learned with the limitation that a fraction of the dataset has to be known to have clean labels. Instead of estimating the noise model directly, the reliability of labels could be estimated instead so that examples with unreliable labels are reweighted to contribute little to the loss function. This was done in \cite{mirikharaji2019learning} by filtering for examples for which gradient directions during training differ greatly from those measured on known clean examples. A similar estimate can be made without requiring clean examples by giving a low weight to examples that tend to produce higher error during training \cite{karimi2020deep}. Alternatively, model predictions \cite{karimi2020deep} (especially those deemed confident by an adversarially trained discriminator \cite{nie2018asdnet}) can be used as pseudo-labels in further training iterations.

The choice of objective function also affects robustness to label noise. Mean absolute error (MAE) in particular exhibits some theoretically grounded robustness to noise~\cite{ghosh2017robust} and is the inspiration for a modified Dice loss that makes Dice less like a weighted mean squared error objective and more like MAE.

True labels can be estimated from multiple imperfect reference labels with expectation minimization (EM)~\cite{dawid1979maximum,whitehill2009whose,welinder2010multidimensional,zhou2015regularized}. Creating multiple segmentation annotations is typically too expensive but lower quality results can be obtained with crowdsourcing~
\cite{orting2019survey}.
Indeed, it has been demonstrated that the most efficient labeling strategy is to collect one high quality label per example for many examples and then estimate the true labels with model-bootstrapped EM~\cite{khetan2017learning}. The authors state that this is effective when ``the learner is robust to noise" and then assume that label errors are random and uniform. This raises the question: which sorts of errors in data labels are deep neural networks robust to?

We show that recent supervised and semi-supervised deep neural network based segmentation models are robust to random ``unbiased'' annotation errors and are much more affected by ``biased'' errors. We refer to errors as biased when the perturbation applied to reference annotations during training is consistent. We test recent supervised and semi-supervised segmentation models, including ``GenSeg''~\cite{vorontsov2019towards}, trained on artificially noisy data with different degrees of bias. Overall, we demonstrate that:

\begin{enumerate}
    \item All models have robustness to unbiased errors.
    \item All models are sensitive to biased errors.
    \item GenSeg is less sensitive to biased errors.
\end{enumerate}

\section{Segmentation models}

Four different deep convolutional neural networks are tested for robustness to label noise in this work. Network architectures and training are detailed in~\cite{vorontsov2019towards}; the models are briefly described below.

\noindent \textbf{Supervised.}
The basic segmentation network is a fully convolutional network (FCN) with long skip connections from an image encoder to a segmentation decoder, similar to the U-Net~\cite{ronneberger2015u}. It is constructed as in~\cite{vorontsov2019towards}, with compressed long skip connections, and trained fully supervised with the soft Dice loss.

\noindent \textbf{Autoencoder.}
The supervised FCN is extended to semi-supervised training by adding a second decoder that reconstructs the input image, as in~\cite{vorontsov2019towards}.

\noindent \textbf{Mean teacher.}
The supervised FCN is extended to mean teacher training as in~\cite{perone2018deep}. A teacher network maintains an exponential moving average of the weights in a student network. When an input has no reference annotation to train on, the student network learns to match the teacher. A potential limitation of this method is that the reliability of the teacher may depend on the size and richness of the annotated training dataset. The supervised FCN architecture is re-used and hyperparameters were selected as in~\cite{vorontsov2019towards}.

\noindent \textbf{GenSeg.}
GenSeg extends the FCN for tumour segmentation by using image-to-image translation between ``healthy'' and ``diseased'' image domains as an unsupervised surrogate objective for segmentation~\cite{vorontsov2019towards}. In order to make a diseased image healthy or \textit{vice versa}, the model must learn to disentangle the tumour from the rest of the image. This disentangling is crucial for a segmentation objective. Importantly, this generative method can learn the locations, shapes, and appearance of tumours conditioned on healthy tissue without relying on tumour annotations.

\section{Model performance on corrupted labels}

We aim to evaluate the robustness of segmentation models to errors in the reference annotations. To that end, we test different types of perturbations applied to the annotations of the training data. Each perturbation (except for permutation) is applied on the fly---that is, the annotation is perturbed from its reference state each time it is loaded during training (once per epoch). Experiments on these various types of perturbation are presented below, along with some loose intuition on their level of bias.

All experiments were performed on the 2D brain tumour dataset proposed in~\cite{vorontsov2019towards}, using the same training, validation, and testing data split. Created from axial slices extracted from the MRI volumes of the 3D brain tumor segmentation challenge (BraTS) 2017 data~\cite{bakas2018identifying}, this dataset includes a set of 8475 healthy slices (without tumour) and a set of 7729 diseased slices (with tumour).

Segmentation performance was evaluated as a Dice score computed on the test set over all inputs combined together. That is, all reference and predicted annotations are stacked together before computing this overlap metric.

\subsection{Random warp}

\begin{figure}[htb!]
    \centering
    \begin{subfigure}{0.135\linewidth}
        \centering
        \includegraphics[width=0.98\linewidth]{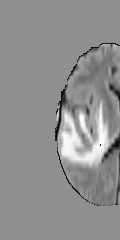}
        \caption{FLAIR}
    \end{subfigure}
    \begin{subfigure}{0.135\linewidth}
        \centering
        \includegraphics[width=0.98\linewidth]{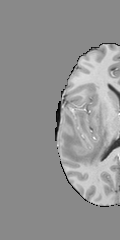}
        \caption{T1}
    \end{subfigure}
    \begin{subfigure}{0.135\linewidth}
        \centering
        \includegraphics[width=0.98\linewidth]{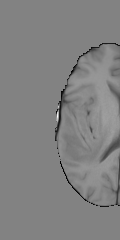}
        \caption{T1C}
    \end{subfigure}
    \begin{subfigure}{0.135\linewidth}
        \centering
        \includegraphics[width=0.98\linewidth]{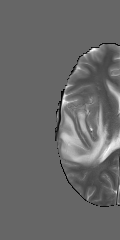}
        \caption{T2}
    \end{subfigure}
    \begin{subfigure}{0.135\linewidth}
        \centering
        \includegraphics[width=0.98\linewidth]{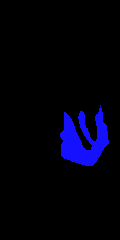}
        \caption{$\sigma=0$}
    \end{subfigure}
    \begin{subfigure}{0.135\linewidth}
        \centering
        \includegraphics[width=0.98\linewidth]{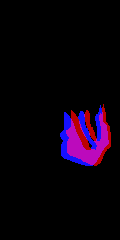}
        \caption{$\sigma=5$}
    \end{subfigure}
    \begin{subfigure}{0.135\linewidth}
        \centering
        \includegraphics[width=0.98\linewidth]{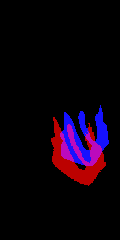}
        \caption{$\sigma=20$}
    \end{subfigure}
    \caption{Example of warped annotations with different $\sigma$ (red) vs original (blue). FLAIR, T1, T1C, and T2 are the MRI acquisition sequences that compose the four channels of each input image.}
    \label{fig:spline_warp}
\end{figure}

We introduced random errors into annotations during training by randomly warping every tumour mask by an elastic deformation. The deformation map was computed with b-spline deformation with a 3$\times$3 grid of control points. Each time warping was applied, each control point was randomly shifted from its initial grid configuration by some number of pixels sampled from a zero-mean Normal distribution with standard deviation $\sigma$. Examples of random warping performed for different values of $\sigma$ are shown in Figure~\ref{fig:spline_warp}.

When warping the tumour mask, the per-pixel label error depends on other pixels, so it cannot be simply averaged out. Nevertheless, there is no bias in the displacement of the control points (all sampled from a zero-mean Normal distribution) and there is no bias in the distribution of shapes produced; in a sense, the original tumour shape remains the average case. Thus, we consider warping to be a largely unbiased error.

\begin{figure}[htb!]
    \centering
    \begin{subfigure}[b]{0.48\linewidth}
        \centering
        \includegraphics[width=0.98\linewidth]{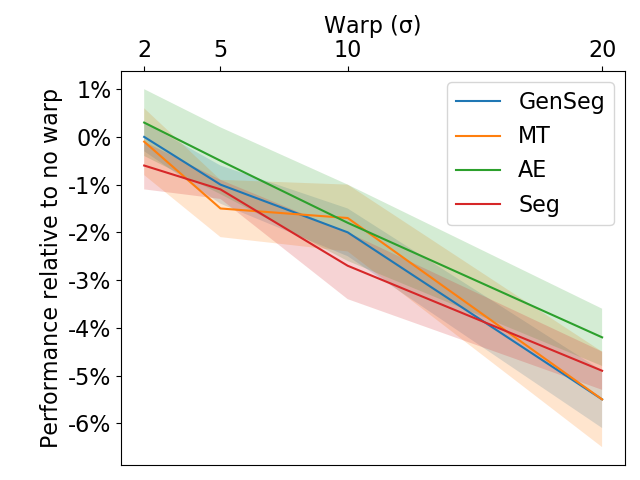}
        \caption{Random warp}
        \label{fig:f1_warp}
    \end{subfigure}
        \begin{subfigure}[b]{0.48\linewidth}
        \centering
        \includegraphics[width=0.98\linewidth]{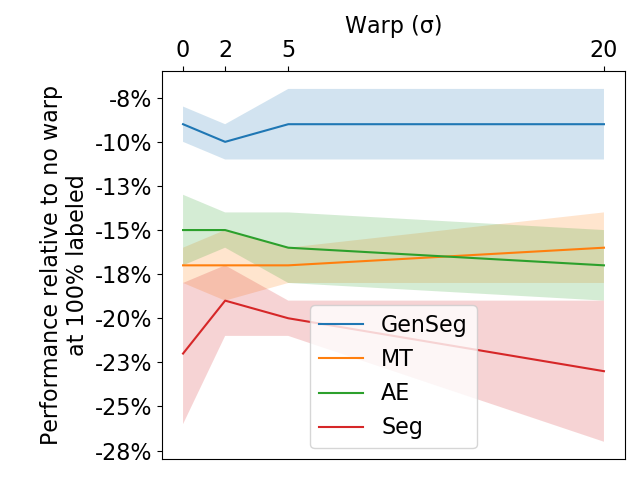}
        \caption{Random warp (1\% labeled)}
        \label{fig:f0_01_warp}
    \end{subfigure}
    \begin{subfigure}[b]{0.48\linewidth}
        \centering
        \includegraphics[width=0.98\linewidth]{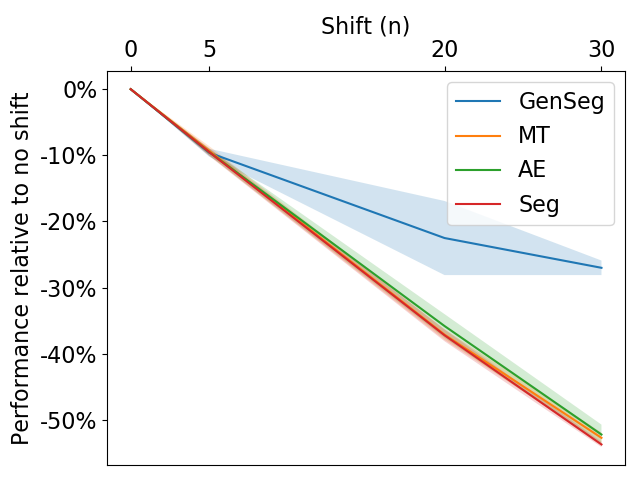}
        \caption{Constant shift}
        \label{fig:f1_shift}
    \end{subfigure}
    \begin{subfigure}[b]{0.48\linewidth}
        \centering
        \includegraphics[width=0.98\linewidth]{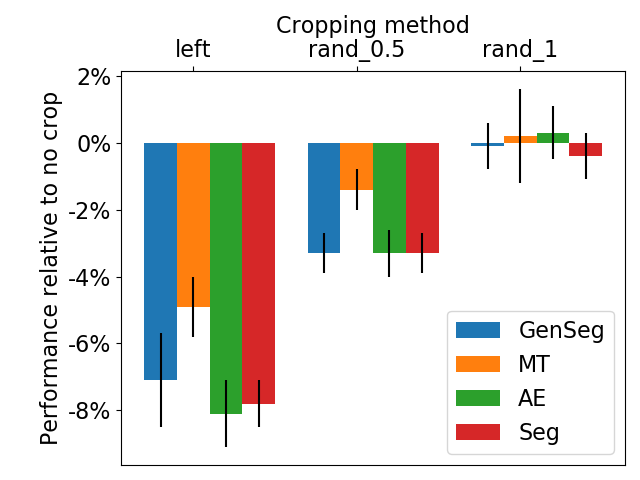}
        \caption{Random crop}
        \label{fig:f1_crop}
    \end{subfigure}
    \caption{Relative segmentation performance for different kinds of errors applied to tumour annotations during training. Compares semi-supervised autoencoding (AE), mean teacher (MT), and GenSeg models, as well as a fully supervised (Seg) network. Performance is relative to each model's peak performance (as Dice score) on clean data. Each experiment was repeated three times. Solid lines: mean; shaded regions: standard deviation. Variance calculations considered both the variance of the presented experiments and of results on clean data.}
\end{figure}

The relative performance of segmentation models trained on warped tumour masks is shown in Figure~\ref{fig:f1_warp}. For different $\sigma$, segmentation performance is evaluated relative to the Dice score achieved with no perturbation of the annotations. All models show a linear relationship of percent reduction in Dice score to $\sigma$, with no measurable reduction in performance for small deformations at $\sigma=2$ and a reduction of only between 4\% and 6\% for unrealistically large deformations at $\sigma=20$. Interestingly, when annotations were only provided for about 1\% of the patient cases, $\sigma$ had no effect on model performance (Figure~\ref{fig:f0_01_warp}). As expected, the semi-supervised autoencoding (AE), mean teacher (MT), and especially GenSeg segmentation methods outperformed fully supervised segmentation in this case. These results show that state of the art segmentation models are surprisingly robust to unbiased deformations of the tumour masks.

\subsection{Constant shift}

We introduced consistent errors into annotations during training by shifting the entire tumour mask by $n$ pixels, creating a consistent misalignment between the target segmentation mask and the input image. Because this error is consistent and the correct (original) annotations cannot be inferred from the distribution of corrupted annotations, we refer to shifting as a biased error. As shown in Figure~\ref{fig:f1_shift}, this kind of error strongly affected all models, resulting in about 10\% lower Dice scores when $n=5$. Interestingly, GenSeg showed remarkable robustness to extreme shift errors, compared to other models. GenSeg was the only model to not show a linear relationship between segmentation performance and the amount of shift, $n$; at an urealistic shift of $n=30$, GenSeg showed a 27\% drop in performance compared to the mean teacher (MT), autoencoding (AE), and purely supervised segmentation (Seg) models which each performed about 53\% worse than on error-free annotations. The same trends were observed when 99\% of the annotations were omitted. These experiments suggest that segmentation models are sensitive to biased errors in the annotations.

\subsection{Random crop}

To further test the effect of bias in annotation errors, we devised three variants of cropping errors. In all cases, we made sure that half of each tumour area is cropped out on averaged. First, we performed a simple a consistent crop of the left side of each tumour (``left''). Second, we cropped out a random rectangle with relative edge lengths distributed in $[0.5, 1]$ as a fraction of the tumour's bounding box (``rand\_0.5''); the rectangle was randomly placed completely within the bounding box. Third, we did the same but with relative edge lengths in $[0, 1]$ (``rand\_0''). We consider ``left'' as the most biased error because it consistently removes the same part of each tumour from the mask. Following this reasoning, ``rand\_0.5'' is a biased error because it consistently reduces the tumour area---that is, there is always a hole in the tumour---but it is less biased than ``left'' because there is no part of the tumour that is never shown to models during training. Finally, ``rand\_0'' is the closest to unbiased because it is inconsistent both in which pixels are made incorrect and in how much of the tumour is removed. Indeed, with ``rand\_0'', there is a chance that the tumour annotation is unmodified.

Experimental results with these random cropping strategies are presented in Figure~\ref{fig:f1_crop}. Similarly to what we observed with unbiased warp errors and biased shift errors, segmentation performance relative to Dice score on clean data drops the most for all models with the biased ``left'' cropping strategy. Surprisingly, performance only decreased between 5\% and 8\% when half of each tumour was consistently missing from the training set. Similarly, the less biased ``rand\_0.5'' strategy reduced segmentation performance but less severely. Finally, the fairly unbiased ``rand\_0'' strategy did not result in reduced segmentation performance at all. These results further suggest that segmentation models are robust to unbiased errors but sensitive to biased errors in the annotations.

\subsection{Permutation}

\begin{figure}[htb!]
    \centering
    \includegraphics[width=0.48\linewidth]{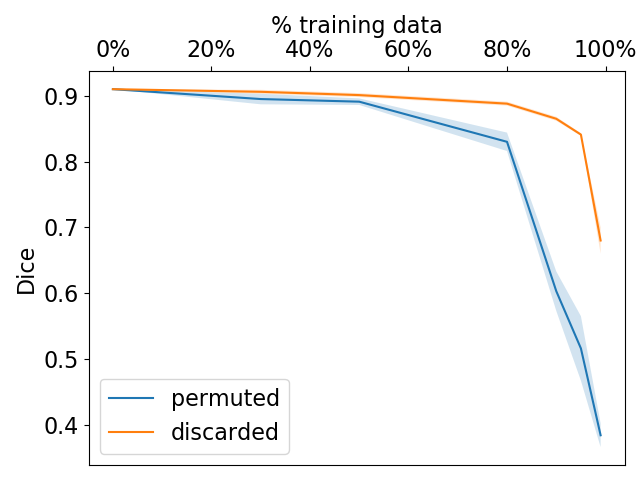}
    \caption{The segmentation Dice score goes down as the percentage of the dataset for which annotations are permuted, or as the percentage of the dataset that is removed from the training set, goes up. Corrupting data by permuting the labels is more detrimental to model performance than discarding the data.}
    \label{fig:f1_permute}
\end{figure}

Occasionally, annotations and images end up being matched incorrectly during the creation of a dataset. We test the effect of this kind of biased error on the performance of a supervised segmentation network. Before training, we randomly permuted the annotations for a percentage of the data. Permutation was done only one time and maintained throughout training. We compared segmentation performance, as the Dice score, to training on clean data but with the same percentage of data discarded from the training dataset. As shown in Figure~\ref{fig:f1_permute}, permutation errors reduce segmentation performance far more than if the corrupted data were simply discarded.

\section{Limitations and future work}

We presented the effects of various biased and unbiased errors, applied to the annotation maps in the training subset of data, on segmentation performance. Although it appears that models are more sensitive to biased errors than unbiased ones, it would be prudent to test many more strategies for introducing error. One simple test could be randomly switching the class of each pixel, independently. This sort of error is commonly considered in the literature for classification; it would be interesting to measure whether segmentation models are more robust to it due to the contextual information from neighbouring pixels.

All tested models used the soft Dice objective since they were trained as in~\cite{vorontsov2019towards}. However, different objective functions have different robustness to label noise~\cite{wang2020noise,ghosh2017robust} so it would be prudent to explore these options further. Furthermore, it would be interesting to evaluate how much of the errors that we introduced could be removed with model-bootstrapped EM~\cite{khetan2017learning} or accounted for with an explicit model of the expected noise~\cite{patrini2017making,sukhbaatar2014training}. Finally, it would be interesting to estimate how much of the error in intra- or inter-annotator variability in medical imaging is systematic biased error, and thus potentially difficult to reduce.

\section{Conclusion}

State of the art deep neural networks for medical image segmentation have some inherent robustness to label noise. We find empirically that while they are robust or partially robust to unbiased errors, they are however sensitive to biased errors. We loosely define biased errors as those which most consistently modify parts of an annotation. We conclude then that when considering on annotation quality (eg. crowdsourcing vs expert annotation) or when working on robustness to label noise, it is particularly important to identify and mitigate against biased errors.

\FloatBarrier
{\small
\bibliographystyle{IEEEtran}
\bibliography{biblio}
}
\end{document}